\documentclass[10pt, conference]{IEEEtran}
\IEEEoverridecommandlockouts
\usepackage{amsmath,amssymb,amsfonts}
\usepackage{algorithmic}
\usepackage{graphicx}
\usepackage{textcomp}
\usepackage{soul}
\usepackage{pdfpages}
\usepackage{xcolor}

\usepackage[utf8]{inputenc}
\usepackage{enumitem}
\usepackage{tikz}
\usetikzlibrary{patterns}
\usepackage{colortbl}
\usepackage{pgfplots}
\usepackage{pgfplotstable}
\pgfplotsset{compat=1.17}

\def\BibTeX{{\rm B\kern-.05em{\sc i\kern-.025em b}\kern-.08em
    T\kern-.1667em\lower.7ex\hbox{E}\kern-.125emX}}

\usepackage[
backend=biber,
style=ieee,
sorting=ynt
]{biblatex}
\bstctlcite{bold}

\begin{document}

\title{\huge Scalable and Efficient Methods for Uncertainty Estimation and Reduction in Deep Learning\\
{\footnotesize DATE PhD Forum 2024}\\[-1.0ex]
}


\author{\IEEEauthorblockN{Soyed Tuhin Ahmed (soyed.ahmed@kit.edu,  Phone: +49 721 608 44859), Thesis Advisor: Prof. Mehdi Tahoori}
\IEEEauthorblockA{\textit{Chair of Dependable Nano Computing (CDNC), Karlsruhe Institute of Technology (KIT), Karlsruhe, Germany} \\
}
\vspace{-2. em}
}

\maketitle



\section{Motivation and Problem Definition}\label{sec:motivation}


Neural networks (NNs) are increasingly being deployed in many critical automated decision-making systems, where functional safety is paramount. 
However, the deployment of NN on resource-constrained safety-critical systems is challenging. Consequently, several hardware accelerator architectures, particularly neuro-inspired \emph{Computation-in-memory (CIM)} with emerging resistive non-volatile memories (memristors), have been proposed~\cite{reuther2021ai}. CIM offers dense on-chip weight storage (4–12 F$^2$ per bit cell) and facilitates efficient parallel computation with significantly reduced power consumption~\cite{ yu2017neuro}. 



Despite this, the predictions made by NN can be \emph{unreliable and uncertain}~\cite{amodei2016concrete}.
It is not guaranteed that they will receive an input that is from the same distribution as the training data, a problem known as out-of-distribution (OOD) uncertainty. Furthermore, hardware non-idealities, such as manufacturing and infield faults, defects, and variations
can change parameters, and activations of CIM implemented NN~\cite{degraeve2015causes}. Due to these non-idealities, another layer of uncertainty is added to the system, which can lead to unpredictable behavior and a reduction in inference accuracy. Therefore, quantifying and mitigating uncertainties is crucial, particularly in safety-critical applications, i.e., by implementing Bayesian NNs (BayNNs) in CIM, proper testing, and fault-tolerance methods. 

However, there are several challenges involved.
Key testing challenges include minimizing testing overhead via test vectors compaction, ensuring non-invasive test generation, and treating the NN as a black box, as many users would consider the trained NN as intellectual property (IP), such as in the case of pre-trained models from Machine Learning as a Service (MLaaS). In contrast, BayNNs can inherently provide uncertainty in prediction, but they are more resource-intensive than conventional NNs. Implementing BayNN algorithms in CIM architectures can mitigate their inherent costs but poses several challenges, such as 
a) implementing parameter distributions in CIM, b) efficiently sampling from it for the inference step, and c) the memory consumption~\cite{ahmed_spinbayes_2023}. 
These bottlenecks lead to the resource scalability issue, where resources such as memory consumption and the number of dropout modules increase with the model's size.

\emph{The objective of this thesis is to overcome the challenges mentioned above using scalable and low-cost methods. Specifically, we explore algorithm-hardware co-design-based solutions to improve the testability, reliability, performance, manufacturing yield, and efficiency of CIM-implemented NNs. We proposed problem-aware training algorithms, novel NN topologies, layers such as normalization, and Dropout, hardware architectures, parameter mapping, and test vector generation methods for our objective~\cite{ahmed2022spindrop, soyed_TNANO23, soyed_nanoarch22, ahmed2023scalable, ahmed_spinbayes_2023, ahmed2023scale, ahmed2022concurrent, ahmed2022single, itc2022, ahmed2022fault, ahmed2022ETS23, ahmed2024inverted, soyed22JETC, ahmed2022process, ahmed2021neuroscrub, ahmed2022neuroscrub, soyed22D_T, ahmed2024VTS} } (\textbf{three best paper nominations)}.


\section{Solutions for uncertainty estimation in CIM}
In this thesis, we explore popular BayNN 
methods, such as Dropout, and variational inference (VI)
for CIM implementation.

\subsubsection{\ul{\textbf{Dropout based BayNNs}}}
In our work~\cite{soyed_nanoarch22, ahmed2022spindrop}, we introduced for \textbf{the first-time dropout-based binary BayNNs}, leveraging stochastic and deterministic properties of spintronic devices for dropout modules and BayNN weight storage, respectively. Thus, existing memristive crossbar arrays can be reused while only modifying peripheral circuits to provide each neuron with a dropout module. Evaluation of the proposed approach shows up to $100\%$ OOD data detection, $\sim 2\%$ increase in accuracy, and $15\%$ increase in accuracy for corrupted data.

Consequently, expanding on the work~\cite{soyed_nanoarch22, ahmed2022spindrop}, we developed the MC-SpatialDropout based BayNN~\cite{soyed_TNANO23} with a specifically designed CIM architecture. In this approach, spatial dropout, which drops entire feature maps of the convolutional layer, is used instead of the conventional dropout module that randomly disables neurons. Consequently, the complexity of the circuit design for implementing the dropout module is reduced. 
Our overall algorithm and hardware co-design lead to a reduction in dropout modules per network by a factor of $9\times$ and energy consumption by $94.11\times$, while maintaining comparable predictive performance and uncertainty estimates.

To further improve the efficiency of BayNN, we introduced a novel dropout approach called \emph{scale dropout}~\cite{ahmed2023scale}. The proposed scale dropout and BayNN learning algorithm are particularly optimized for CIM architectures and require \textbf{only a single dropout module for the entire model}. Furthermore, a \emph{layer-dependent adaptive scale dropout} approach is proposed that mitigates extensive design space exploration to find the optimal dropout probability and location of the dropout layer for each model. Consequently, our approach improves inference accuracy by up to $1\%$, can detect OOD by up to $100\%$, and provides high-quality uncertainty estimates, while achieving more than $100\times$ energy savings compared to existing methods.

\subsubsection{\ul{\textbf{VI Based BayNNs}}}

Unlike traditional VI-based BayNNs, in our work~\cite{ahmed2023scalable}, we introduce a novel Bayesian NN framework and CiM architecture where Bayesian treatment is only applied to the small parameter group, e.g., scale vector. Larger parameter groups (e.g., weights) are kept deterministic (single-point value), leading to \textbf{first binary VI-based BayNNs framework with spintronic-based CIM implementation}. Ultimately, the proposed approach requires up to $70\times$ and $158.7\times$ lower power consumption and memory requirements, respectively, compared to traditional methods while maintaining comparable inference accuracy and uncertainty estimates.


To further improve the accuracy and efficiency of BayNN for \emph{harder tasks}, in our work~\cite{ahmed_spinbayes_2023}, we introduce \emph{SpinBayes}, a novel BayNN topology, and the \emph{Bayesian in-memory approximation}. 
The proposed approaches allow efficient mapping and sampling of BayNN distributions in the CiM architecture. 
Our methods lead to a significant enhancement in performance: classification accuracy improves by up to $1.14\%$, uncertainty estimation by $20.16\%$, and OOD data detection reaches $100\%$. Moreover, our approach results in up to $80\times$ energy consumption and $8\times$ memory overhead reduction. In particular, our approach maintains resource scalability, requiring only three RNGs irrespective of model size.

As a result, we are able to solve all the challenges discussed in Section~\ref{sec:motivation} while achieving high efficiency and accuracy. 


\section{Solutions For Testing CIM Architectures}




To achieve our goal of compacting the functional test vectors that meet our requirements, we introduce the \emph{approximate gradient ranking method}~\cite{itc2022}. Our approach exploits the varying difficulty of NN inputs during training, identifying those inputs requiring more parameter tuning for self-testing NNs, as they are more sensitive to parameter changes. Our approach achieves up to $100\%$ test coverage using only a maximum of $0.128\%$ of training data as test vectors, a substantial reduction compared to existing methods. Additionally, our approach can detect hard-to-detect faults with high test coverage.



Furthermore, to reduce the testing overhead of \emph{large-scale NN models} to an absolute minimum, we introduce the \textbf{one-shot} testing method that requires only a \textbf{single test vector and a forward pass}, as detailed in~\cite{ahmed2022single}. 
Our approach is based on the output distribution shift in NNs. An optimization algorithm is proposed that tunes the one-shot test vector through the gradient descent algorithm to achieve a unit Gaussian output distribution. By analyzing distribution shift (compared to unit Gaussian), we can test NN models as big as $200$ layers and $44.5\times 10^6$ parameters while consistently achieving $100\%$ test coverage in a \emph{single shot} with $\sim 10000\times$ lower memory overhead and MAC operations.


To detect soft or transient faults while the NN is computing its prediction, our work ~\cite{ahmed2022concurrent} introduces a novel NN topology for concurrent self-testing without extra forward passes or test vector storage. Our method \emph{fingerprints} the fault-free status of the NN using a specialized training approach, allowing real-time matching of this fingerprint online to test the NN up to $100\%$ test coverage, low false positive rates, and maintained similar primary task performance as the baseline. Compared to existing methods, our approach reduces memory overhead by up to $243.7$ MB, MAC operations by approximately $10000\times$, and false positive rates by up to $89\%$.

\section{Solutions for improving Reliability of CIM}

In work~\cite{ahmed2022neuroscrub}\cite{ahmed2021neuroscrub}, to mitigate data retention faults and the aging-induced drift problem, we propose a novel approximate scrubbing technique with virtually zero storage overhead. The proposed approach is equipped with a specifically designed NN training algorithm and a one-time mapping technique for CIM. Consequently, the proposed method can improve the inference accuracy by $87.76\%$ over the device operation time.



In work \cite{ahmed2022fault, soyed22D_T}, we propose the post-manufacturing and in-field re-calibration method to mitigate the impact of manufacturing variations and defects. We proposed a functional automatic test pattern generation (ATPG) method and the approximate batch normalization (ApproxBN) method that requires only $0.2\%$ of training data for re-calibration but can still regain inference accuracy by up to $72.32\%$ and manufacturing yield by up to $100\%$. This work showed for the first time that the re-calibration method can be applied to improve the reliability of binary NNs and mitigate manufacturing defects.




In work~\cite{ahmed2022process, soyed22JETC}, we propose a variation-aware NN training algorithm and a design-time (post-deployment) reference generation method to improve the robustness 
under different thermal and process variation scenarios (up to $125^\circ$C temperature). The work also focuses on binarizing partial currents to increase the sensing margin and can improve inference accuracy by up to $20.51\%$. Hence, analog-to-digital (ADC) converter-less inference is feasible in CIM.



In work~\cite{ahmed2024VTS}, we propose zero overhead Error correction codes (ECCs)-based (guaranteed) fault tolerance for a certain number of soft faults. Conventional ECC has as much as 15\% memory overhead. Our approach integrates parity bits of ECCs into NN parameters using a multi-task NN learning algorithm and a specifically designed ECC, while maintaining the same number of parameters and accuracy. 

Furthermore, in our work~\cite{ahmed2022ETS23} we proposed an approach to ensure the continuous availability of neuromorphic hardware in the event of system downtime due to failures. System downtime is inevitable, but it is not acceptable in many applications. Our work aims to complement time-consuming testing or fault tolerance methods, such as retraining for always-on NN applications. We propose to create local approximations of only important modules (rather than full copies) of the NN based on the design-time sensitivity analysis. Our approach reduces hardware overhead by up to $99.37\%$ compared to conventional redundancy-based approaches.



Lastly, in work~\cite{ahmed2024inverted}, we proposed \textbf{a self-healing BayNN} based on inverted normalization with affine Dropout, a novel normalization and dropout method to enhance the robustness and inference accuracy under memristive non-idealities.
The proposed approach improves inference accuracy by up to 55.62\%, the root mean square error (RMSE) score LSTM-based time series prediction is reduced by up to 46.7\% while being able to detect 78.95\% of OOD data (an improvement of 14.61\% compared to previous work).


\vspace{-0.2cm}
\def\bibfont{\footnotesize}
\def\bibstyle{IEEEtran}
\footnotesize{\printbibliography[keyword={ref},title={References}]}
\footnotesize{\printbibliography[keyword={soyed}, title={Publications (1st Authored )}]}

@article{degraeve2015causes,
  title={Causes and consequences of the stochastic aspect of filamentary RRAM},
  author={Degraeve, Robin and others},
  journal={Microelectronic Engineering},
  volume={147},
  pages={171--175},
  year={2015},
  publisher={Elsevier}
}

@inproceedings{reuther2021ai,
  title={AI accelerator survey and trends},
  author={Reuther, Albert and others},
  booktitle={2021 IEEE (HPEC)},
  pages={1--9},
  year={2021},
  organization={IEEE}
}

@article{amodei2016concrete,
  title={Concrete problems in AI safety},
  author={Amodei, Dario and Others},
  year={2016},
  journal={arXiv preprint arXiv:1606.06565}
}

@ARTICLE{yu2017neuro,
  author={Yu, Shimeng},
  journal={Proceedings of the IEEE}, 
  title={Neuro-inspired computing with emerging nonvolatile memorys}, 
  year={2018},
  volume={106},
  number={2},
  pages={260-285}}

@ARTICLE{ahmed2022spindrop,
  author={Ahmed, Soyed Tuhin and Danouchi, Kamal and Münch, Christopher and Prenat, Guillaume and Anghel, Lorena and Tahoori, Mehdi B.},
  journal={IEEE Journal on Emerging and Selected Topics in Circuits and Systems}, 
  title={SpinDrop: Dropout-Based Bayesian Binary Neural Networks With Spintronic Implementation}, 
  year={2023},
  volume={13},
  number={1},
  pages={150-164},
  doi={10.1109/JETCAS.2023.3242146}
}

@inproceedings{soyed_nanoarch22,
  title={Binary bayesian neural networks for efficient uncertainty estimation leveraging inherent stochasticity of spintronic devices},
  author={Ahmed, Soyed Tuhin and others},
  booktitle={NANOARCH'22: 17th ACM International Symposium on Nanoscale Architectures},
  pages={1--6, \textbf{(Best paper candidate)}},
  year={2022},
  organization={ACM}
}

@article{soyed_TNANO23,
  title={Spatial-SpinDrop: Spatial Dropout-based Binary Bayesian Neural Network with Spintronics Implementation},
  author={Ahmed, Soyed Tuhin and Danouchi, Kamal and Hefenbrock, Michael and Prenat, Guillaume and Anghel, Lorena and Tahoori, Mehdi B},
  journal={Under Review at IEEE Transactions on Nanotechnology (TNANO)},
  year={2023},
}

@article{ahmed2023scale,
  title={Scale-Dropout: Estimating Uncertainty in Deep Neural Networks Using Stochastic Scale},
  author={Ahmed, Soyed Tuhin and Danouchi, Kamal and Hefenbrock, Michael and Prenat, Guillaume and Anghel, Lorena and Tahoori, Mehdi B},
  journal={Under Review at IEEE Transactions on Circuits and Systems I: Regular Papers},
  year={2023}
}

@inproceedings{ahmed2023scalable,
  title={Scalable Spintronics-based Bayesian Neural Network for Uncertainty Estimation},
  author={Ahmed, Soyed Tuhin and Danouchi, Kamal and Hefenbrock, Michael and Prenat, Guillaume and Anghel, Lorena and Tahoori, Mehdi B},
  booktitle={2023 Design, Automation \& Test in Europe Conference \& Exhibition (DATE)},
  pages={1--6},
  year={2023},
  organization={IEEE}
}

@article{ahmed_spinbayes_2023,
	title = {{SpinBayes}: {Algorithm}-{Hardware} {Co}-{Design} for {Uncertainty} {Estimation} {Using} {Bayesian} {In}-{Memory} {Approximation} on {Spintronic}-{Based} {Architectures}},
	volume = {22},
	issn = {1539-9087},
	shorttitle = {{SpinBayes}},
	url = {https://doi.org/10.1145/3609116},
	doi = {10.1145/3609116},number = {5s},
	urldate = {2023-09-13},
	journal = {ACM Transactions on Embedded Computing Systems},
	author = {Ahmed, Soyed Tuhin and Danouchi, Kamal and Hefenbrock, Michael and Prenat, Guillaume and Anghel, Lorena and Tahoori, Mehdi B.},
	month = sep,
	year = {2023}
}

@inproceedings{ahmed2022concurrent,
  title={Concurrent Self-testing of Neural Networks},
  author={Ahmed, Soyed Tuhin and Others},
  booktitle={Under review at 2024 IEEE 42th VLSI Test Symposium (VTS)},
  pages={1--7},
  year={2024},
  organization={IEEE}
}

@article{ahmed2022single,
  title={One-Shot Online Testing of Deep Neural Networks Based on Distribution Shift Detection},
  author={Ahmed, Soyed Tuhin and Others},
  journal={Under Review at IEEE Transactions on Computer-Aided Design of Integrated Circuits and Systems (TCAD)},
  year={2023},
  publisher={IEEE}
}

@INPROCEEDINGS{itc2022,
  author={Ahmed, Soyed Tuhin and others},
  booktitle={2022 IEEE International Test Conference (ITC)}, 
  title={Compact Functional Test Generation for Memristive Deep Learning Implementations using Approximate Gradient Ranking}, 
  year={2022},
  volume={},
  number={},
  pages={},
  doi={}}

@article{ahmed2022neuroscrub,
  title={NeuroScrub+: Mitigating Retention Faults Using Flexible Approximate Scrubbing in Neuromorphic Fabric Based on Resistive Memories},
  author={Ahmed, Soyed Tuhin and Others},
  journal={IEEE Transactions on Computer-Aided Design of Integrated Circuits and Systems (TCAD)},
  year={2022},
  publisher={IEEE}
}

@inproceedings{ahmed2021neuroscrub,
  title={Neuroscrub: Mitigating retention failures using approximate scrubbing in neuromorphic fabric based on resistive memories},
  author={Ahmed, Soyed Tuhin and Others},
  booktitle={IEEE IEEE European Test Symposium (ETS)},
  pages={},
  year={2021},
  organization={}
}

@inproceedings{ahmed2022fault,
  title={Fault-tolerant Neuromorphic Computing with Functional ATPG for Post-manufacturing Re-calibration},
  author={Ahmed, Soyed Tuhin and Others},
  booktitle={IEEE 40th VLSI Test Symposium (VTS)},
  pages={1--7, \textbf{(Best paper candidate)}},
  year={2022},
  organization={IEEE}
}

@ARTICLE{soyed22D_T,
  author={Ahmed, Soyed Tuhin and Tahoori, Mehdi B.},
  journal={IEEE Design \& Test}, 
  title={Fault-Tolerant Neuromorphic Computing With Memristors Using Functional ATPG for Efficient Recalibration}, 
  year={2023},
  volume={40},
  number={4},
  pages={42-50},
  doi={10.1109/MDAT.2023.3270126}}

@inproceedings{ahmed2022process,
  title={Process and Runtime Variation Robustness for Spintronic-Based Neuromorphic Fabric},
  author={Ahmed, Soyed Tuhin and others},
  booktitle={2022 IEEE European Test Symposium (ETS)},
  pages={},
  year={2022},
  organization={IEEE}
}

@article{soyed22JETC,
author = {Ahmed, Soyed Tuhin and others},
title = {Design-Time Reference Current Generation for Robust Spintronic-Based Neuromorphic Architecture},
year = {2023},
issue_date = {January 2024},
publisher = {Association for Computing Machinery},
address = {New York, NY, USA},
volume = {20},
number = {1},
issn = {1550-4832},
doi = {10.1145/3625556},
journal = {J. Emerg. Technol. Comput. Syst.},
month = {11},
articleno = {2},
numpages = {20}
}

@inproceedings{ahmed2022ETS23,
  title={Online Fault-Tolerance for Memristive Neuromorphic Fabric Based on Local Approximation},
  author={Ahmed, Soyed Tuhin and Rakhmatullin, Roman and Tahoori, Mehdi B},
  booktitle={2023 IEEE European Test Symposium (ETS)},
  pages={1--4},
  year={2023},
  organization={IEEE}
}

@inproceedings{ahmed2024VTS,
  title={Embedding Error Correction Codes in
Neural Network Weight using Multi-task
Learning},
  author={Ahmed, Soyed Tuhin and Others},
  booktitle={Under review at 2024 IEEE 42th VLSI Test Symposium (VTS)},
  pages={1--7},
  year={2024},
  organization={IEEE}
}

@inproceedings{ahmed2024inverted,
  title={Enhancing Reliability of Neural Networks at the Edge: Inverted Normalization with Stochastic Affine Transformations},
  author={Ahmed, Soyed Tuhin and others},
  booktitle={2024 Design, Automation \& Test in Europe Conference \& Exhibition (DATE)},
  pages={1--6, \textbf{(Best paper candidate)}},
  year={2024},
  organization={IEEE}
}

\end{document}